\documentclass{svproc}
\usepackage{url}
\usepackage{graphicx}
\usepackage{multicol}
\usepackage{footmisc}
\usepackage{cite}
\usepackage{amsmath,amssymb,amsfonts}
\usepackage{algorithmic}
\usepackage{textcomp}
\usepackage{xcolor}

\begin{document}
\mainmatter              
\title{Large-Scale Data Parallelization of Product Quantization and Inverted Indexing Using Dask}
\titlerunning{Large-Scale Data Parallelization of PQ and II Using Dask}
\author{Ashley N. Abraham\inst{1} \and Andrew Strelzoff\inst{1} \and
Haley Dozier\inst{1} \and Althea Henslee\inst{1} \and Mark A. Chappell\inst{2}}
\authorrunning{Ashley N. Abraham et al.} 
%
\tocauthor{Ashley N. Abraham, Andrew Strelzoff, Haley Dozier, Althea Henslee, and Mark A. Chappell}
\institute{Information Technology Laboratory, U.S Army Engineer Research and Development Center, Vicksburg, MS, USA\\
\email{ Ashley.N.Abraham@usace.army.mil},
\texttt{https://www.erdc.usace.army.mil/Locations/ITL/}
\and
Environmental Laboratory, U.S Army Engineer Research and Development Center, Vicksburg, MS, USA}
\maketitle
\begin{abstract}
Large-scale Nearest Neighbor (NN) search, though widely utilized in the similarity search field, remains challenged by the computational limitations inherent in processing large scale data.  In an effort to decrease the computational expense needed, Approximate Nearest Neighbor (ANN) search is often used in applications that do not require the exact similarity search, but instead can rely on an approximation.  Product Quantization (PQ) is a memory-efficient ANN effective for clustering all sizes of datasets. Clustering large-scale, high dimensional data requires a heavy computational expense, in both memory-cost and execution time. This work focuses on a unique way to divide and conquer the large scale data in Python using PQ, Inverted Indexing and Dask, combining the results without compromising the accuracy and reducing computational requirements to the level required when using medium-scale data.
\keywords{
Machine Learning, Big Data, Approximate Nearest Neighbor Search, Product Quantization, Parallelization, 
}
\end{abstract}

\section{Introduction}
Similarity search algorithms have been widely studied due to their broad range of applicability in modern solutions, ranging from autonomous self-driving cars to social media. Despite their broad use, applying similarity search solutions to large data remains challenging. While many existing memory-efficient algorithms and libraries are available to solve small-scale and medium-scale data problems, when it comes to large-scale datasets, many computer systems are not equipped with adequate memory to implement a similarity search-based solution. However, similarity search can easily be implemented on datasets of both small and medium scale using nearest neighbor (NN) algorithms, and with enough memory, even large-scale data can also be processed. Not all applications require the exact NN; some applications are able to utilize Approximate Nearest Neighbor (ANN) algorithms to further reduce computational expense, as ANNs are generally memory-efficient algorithms \cite{FAISS}\cite{Optimized_PQ}\cite{PQSurvey}. Even with ANNs, large-scale data remains challenging, as there is a trade-off between accuracy and memory-efficiency/run-time \cite{PQSurvey}. 
 
In this paper we are utilizing the power of distributed parallel computing to improve the memory-efficiency and reduce run-time required by the large-scale ANN problem. 

\section{Background and Related Work}
\subsection{Product Quantization}
One popular ANN approach is Product Quantization. Product Quantization, (PQ), algorithms generally decompose high-dimensional, large-scale data into lower-dimensional, representational data, which in turn allows for efficiency in similarity search \cite{PQ}. Previous research has already led to many improvements, increasing the memory efficiency while reducing the run-time without compromising the accuracy, from hardware-based SIMD efficiency improvements to local optimization of PQ \cite{Optimized_PQ, nanopq}\cite{PQSurvey}.  

In the fitting process, when given a subspace $M$, and a code size $Ks$, the PQ algorithm takes the given data and separates the dimensions into subspaces of smaller dimensions. Each of the smaller subspaces is then processed by a k-means clustering algorithm and fitted into given $Ks$ number of centroids. These resulting centroids are called codebooks or codewords. During in the encoding process, each subspace goes through a vector quantization, during which each row of subspace data is represented by its assigned subspace centroid so that the entire row of data is represented by M centroids, known as codes \cite{PQ}. PQ also has a function to calculate the distance between each item and its centroid. When querying for a new item, the centroid and distance are calculated, and data points with the same centroid and those that fall within the threshold distance are considered to be similar, or approximately nearest to the query.

\begin{figure}
    \centering
    \includegraphics[width = 10cm]{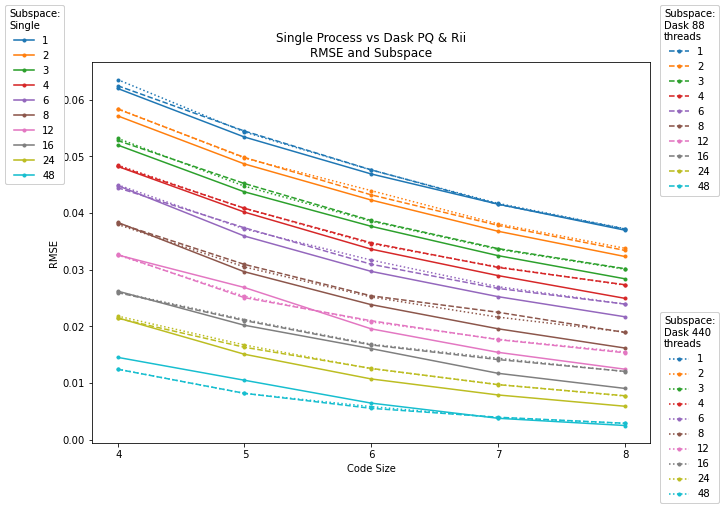}
    \includegraphics[width = 10cm]{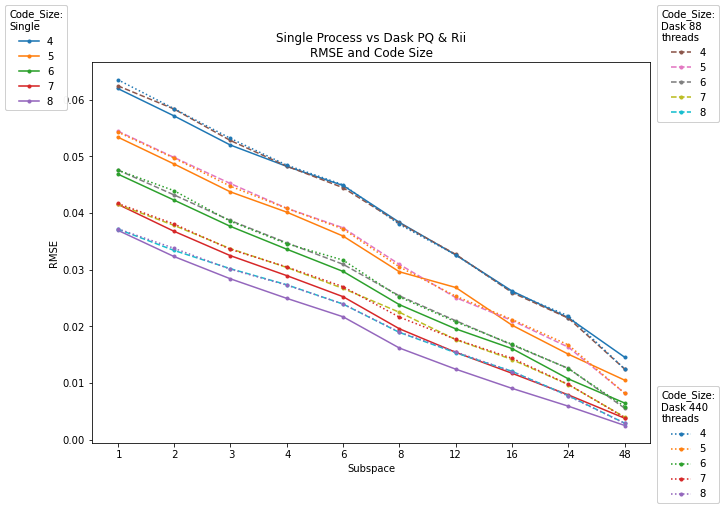}
    \caption{Accuracy trade-offs between Subspace and Code Size: as subspace (top) and code size (bottom) numbers increases the RMSE decreases. Accuracy results using parallelization from Dask PQ and RII RMSE is very close to the Single process}
    \label{fig:IndividualRMSE}
\end{figure}

\begin{figure}
    \centering
    \includegraphics[width = 10cm]{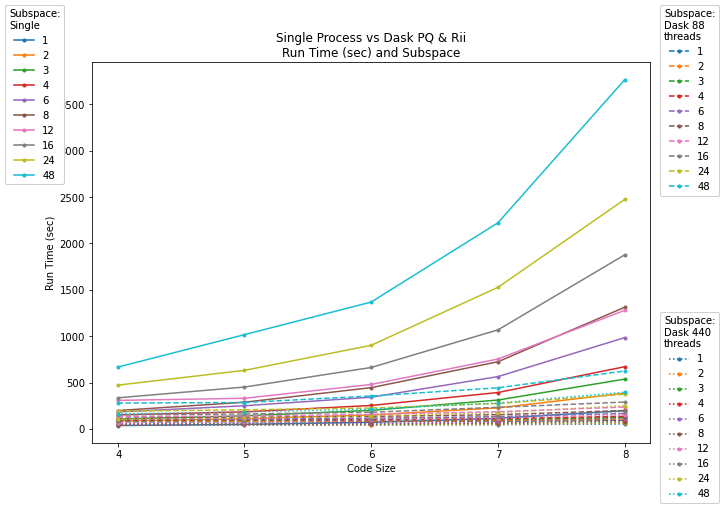}
    \includegraphics[width = 10cm]{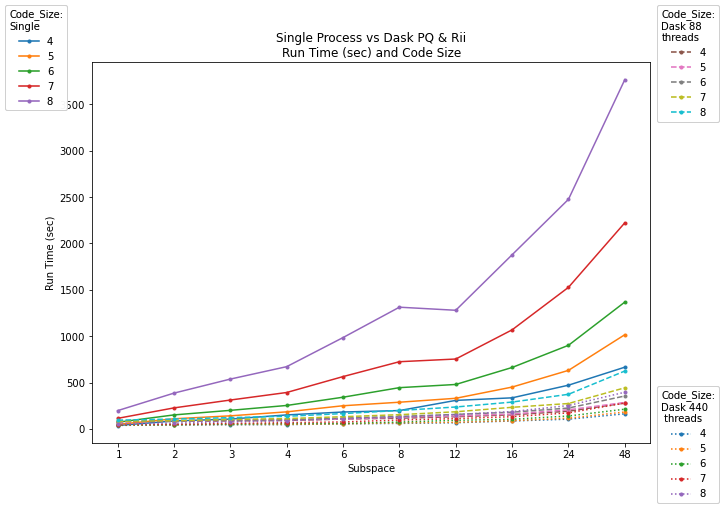}
    \caption{Run Time trade-offs between Subspace and Code Size: as subspace (top) and code size (bottom) increases the Run Time increases significantly in the Single process and the gains are seen by using parallelization from Dask PQ and RII, it is significantly faster than the Single process}
    \label{fig:IndividualRunTime}
\end{figure}

The subspace and code size parameter controls the trade-off of higher accuracy and memory-cost, as the subspace or code size value increases, the rate of error decreases, as seen in figure \ref{fig:IndividualRMSE}, and the run time increases, as see in figure \ref{fig:IndividualRunTime}. For this work, Root Mean Squared Error (RMSE) is used to determine accuracy and execution run time is used to determine the speed and memory-cost.


\subsection{Inverted Indexing}
Although the distance table from PQ can be used to identify the ANNs, the search is not efficient or optimized, and much research has been focused on optimizing the encoded codes and using them to query efficiently. One such approach is using inverted indexing \cite{InvertedIndex}. In this approach, the encoded codes from each row are hashed using LSH \cite{LSH1} \cite{RII}, and the ids are stored. This inverted method of indexing provides a faster lookup of queries of ANN\cite{RII}. The query results not only include the first nearest neighbor, but the second, third, and to the k-th nearest neighbor as well. These results are critical, as in many applications, it is necessary to gather k-th nearest neighbors of the query, not just the first one. 

\subsection{Parallelization of ANN}

While the idea of parallelization is not new, it was not widely utilized until recently when the cloud computing technologies, such as Amazon Web Services, Microsoft Azure, High Performance Computers (HPC) and other technologies, were adopted by the industry. During parallelization, a large problem is broken down into smaller pieces, and solved using multiple computer threads. And by using parallelization, large-scale data is divided and processed individually, and the results are gathered and processed, significantly lowering increasing memory efficiency while lowering run-time execution.

\subsection{Python Packages for Parallelization}

In the past decade, many open source libraries have made data science more easily accessible for researchers of all fields, especially using the Python platform. Python has a huge support for Data Science, AI, Machine Learning, Natural Language Processing, Vision, and other fields. Given the nature of support and Python being the industry standard for data science, this study utilizes Python libraries as well. The programming language is kept consistent across all tools for this study. Three main python packages are utilized: NanoPQ, a 100 percent PQ library writing in pure Python; Reverse Inverted Index (RII), a memory-efficient search library to query the PQ codes for ANN; and Dask, a distributable parallel computing library used to divide and conquer the large-scale data.   

There are other parallelization tools and libraries available to solve the large-scale problem, such as SCOOP, Apache Spark, and others. And while other libraries are available to implement PQ and inverted indexing, such as FAISS \cite{FAISS}, Optimized PQ \cite{Optimized_PQ}, among others, these may be optimized using hardware or other optimizations. This study aims for optimization through the use of parallel computing.

\subsubsection{Facebook AI Similarity Search (FAISS)}
FAISS is developed by Facebook AI Research, the industry standard library developed by the original authors of PQ. FAISS contains several methods to solve the ANN and it contains many parameter tunning options and both CPU/GPU optimizations built into it.

\subsubsection{Scalable COncurrent Operations in Python (SCOOP)}
SCOOP \cite{SCOOP} is a distributed task module allowing concurrent parallel programming on various environments, from heterogeneous grids to supercomputers. 

\subsubsection{Spark}
Apache Spark \cite{Apache_Spark} using Hadoop, is a unified engine that runs from single node to multiple nodes.
There are existing Spark APIs in Python. Apache follows the Map-Shuffle-Reduce paradigm.

\section{Methodology}

\begin{figure}[!ht]
    \centering
    \includegraphics[width=10cm]{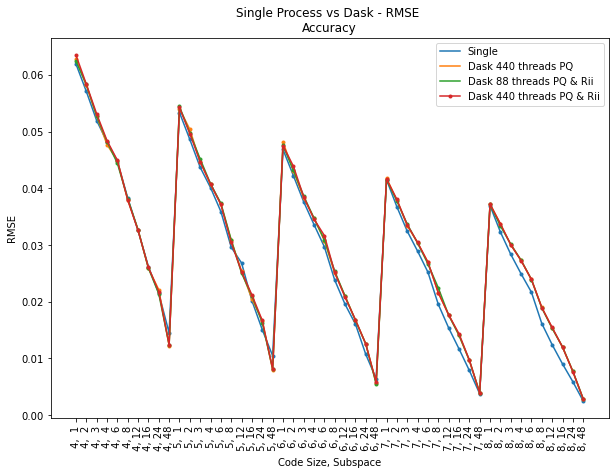}
    \includegraphics[width=10cm]{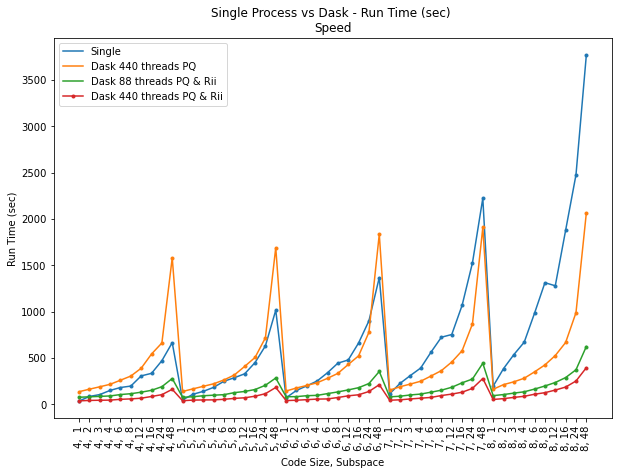}
    \caption{Parallelization accuracy (top) using Dask 88 and 440 threads is comparable to the accuracy from Single process. The run time (bottom) benefits gained from parallelization using Dask PQ and RII (88 or 440 threads) are significant than using just Dask PQ, or the Single process}
    \label{fig:Summary}
\end{figure}

There are multiple possible methodologies to consider when approaching parallel computing: the first is to divide the data row-wise; the second is the divide the data column-wise, and the third is a combination of both. For this study, the data is divided row-wise into chunks, processed in parallel, and the results combined. One challenge when implementing PQ in parallel is the handling of centroid scope when encoding the data. When the chunks are processed using PQ in parallel, encodings are based on that chunk of data, and therefore, the encoded data's scope is limited to that chunk only. The encoded codes from the chunks are returned from the parallel tasks and are later combined together. Moreover, when combining the locally encoded data after parallellization, the representation in the global scope is lost. To overcome this loss and preserve the codes processed in parallel in the global scope, we propose to decode the local centroids and return the values as results. The number of data that is returned is only the centroid's actual row values. All the decoded centroids from the parallel tasks are combined together as a new centroid dataset representing the much large scale dataset, and a new global PQ model is created and trained from the centroid dataset. Using this new global PQ model, the original dataset is encoded and decoded to capture the reconstruction error as Root Mean Square Error to accurately depict the model precision. As shown in the figure \ref{fig:Summary} , the reconstruction error is within a small margin of 1/10th to 1/100th. The main benefit of parallelization comes from decreasing computational expense, allowing memory-efficient, run-time execution savings.

 As an example, assuming the data is 10M points, split into 100 chunks and processed in parallel using PQ, each chunk will contain 100,000 data points. Further, assuming a subspace of 8 is chosen, each row of data will have eight representative centroids after encoding in PQ. 

\begin{figure}
    \centering
    \includegraphics[width=10cm]{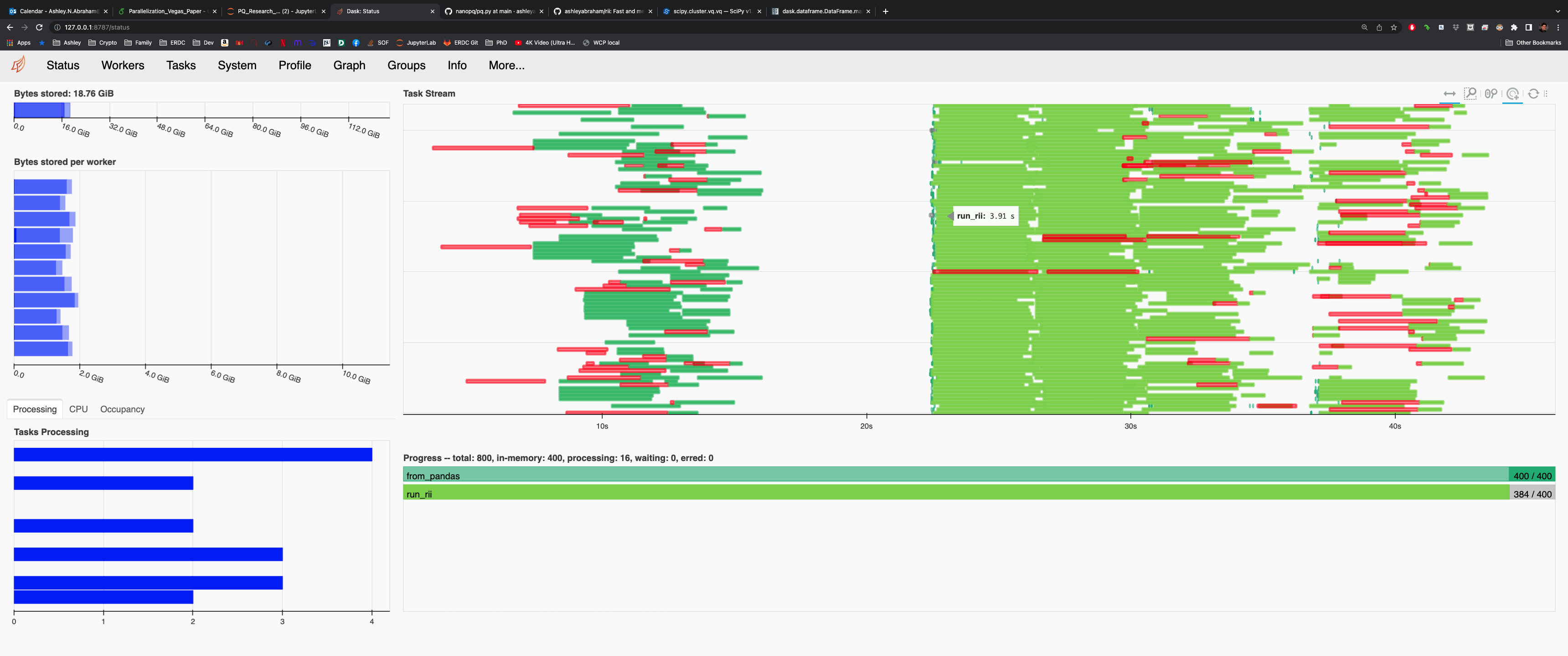}
    \caption{A screenshot of Dask Dashboard, displaying 440 threads processing PQ and RII tasks in parallel.}
    \label{fig:Dask_440threads}
\end{figure}

The data used in this work contains 6.7M rows and 48 columns, sampled from Soil Grids 250 meter\cite{SoilGrids}, the data was split into 400 chunks, 16,750 rows of data in each chunk, and processed parallelly using these case studies. 
Three case studies are included: a single system approach, a single node Dask approach, and a 10 node Dask cluster approach. The Single system did not utilize the multi-threaded parallelization and is included as a base line.
The single node Dask system used in this work contains 88 threads, including 11 workers with each worker managing eight threads. Each system in the 10 node Dask cluster contains 44 threads, combining for a total of 440 threads, including 11 workers in each node, with each worker managing 4 threads.
The single system does not contain any parallelization to it, the other case studies are shown with parallelization only applied to PQ, and finally parallelization done to both PQ and Rii, as show in figure \ref{fig:Summary}

\subsection{Python Libraries used}
This work uses three open-source, Python libraries: NanoPQ, Rii, and Dask.

\subsubsection{NanoPQ}
NanoPQ \cite{nanopq} is a pure python implementation of PQ and Optimized PQ algorithms. Its reconstruction error and accuracy rates are very similar to the popular PQ library called FAISS \cite{FAISS}. However, FAISS has been optimized in many different aspects, and it would be difficult to accurately assess only the benefits of applying parallel computing using it. 

\subsubsection{Rii}
Reconfigurable Inverted Index (Rii) \cite{RII}, an IVFPQ-based \cite{FAISS} fast and memory efficient approximate nearest neighbor search method with a subset-search functionality, utilizes NanoPQ and the encoded codes to build an inverse index table using Locality-sensitive Hashing (LSH) to quickly query and return the nearest neighbors. Rii provides optimized lookup of the nearest neighbors, which also has features to add, reconfigure, and merge new codes to the Rii model. This feature is utilized to create a Rii model in parallel using Dask to provide even faster run-time and memory-efficiency than achievable without it.

\subsubsection{Dask}
The Dask \cite{Dask} Python library provides lower-level and higher parallel APIs than SCOOP or Spark. In this project, these attributes of Dask are utilized to speed up the large-scale creation of PQ and the Rii models. See figure \ref{fig:Dask_440threads} for an example dashboard of parallelization task stream.

\section{Results}

Parallel computing results in a negligible difference in reconstruction error, as parallelized PQ produces an almost identical accuracy as the single processor-based PQ.
Also of note, Parallelization is not suited for all scales of data, as seen in the Figures \ref{fig:IndividualRMSE} and \ref{fig:IndividualRunTime}. Even though the parallel implementation produces equivalent accuracy, its benefit are seen only when used in combination with large-scale data, as shown in graph.

These figures clearly support that PQ should not be implemented with parallel computing in all applications. Instead, PQ for small and medium-scale dataset can perform well without the overhead of parallelization. However, solutions for large-scale data will benefit from parallelization, with significant gains using a single-node with multi-cores implementation, but the highest benefits will be attained with a multi-node and multi-core setup.

\section{Future Work}
Many opportunities exist for further work in this area, including investigation of other tools, HPC use, and other parallel computing approaches. The comparison with other tools such as SCOOP, Apache Spark, and SIMD based in multi-node cluster platforms will allow more insight into further applications. Further study using different chunk sizes, multiple nodes in HPC, with multiple cores and threads, would increase scalability to even larger scale datasets. Then, the column wise approach and combination of row and column wise approach of parallelization can be studied, though it requires modification of the PQ and Inverse Index libraries to use Dask and Dask Dataframe, and to study the fine tunning of PQ and Rii optimizations. And finally, to apply PQ and Rii parallelization to the multi billion scale Soil Grids 250 meter 2017 dataset \cite{SoilGrids}.


\bibliographystyle{splncs04}
%

\end{document}